\documentclass[conference]{IEEEtran}
\IEEEoverridecommandlockouts
\usepackage{cite}
\usepackage{amsmath,amssymb,amsfonts}
\usepackage{graphicx}
\usepackage{textcomp}
\usepackage{xcolor}
\def\BibTeX{{\rm B\kern-.05em{\sc i\kern-.025em b}\kern-.08em
    T\kern-.1667em\lower.7ex\hbox{E}\kern-.125emX}}

\usepackage{amsthm}
\usepackage{amsmath}
\usepackage{bm}
\usepackage{color}
\usepackage{etex}
\usepackage{qtree}
\usepackage{graphicx}
\usepackage{multirow}
\usepackage{multicol}
\usepackage{subfigure}
\usepackage{url}
\usepackage{thmtools}
\usepackage{ctable}
\usepackage{tabularx}
\usepackage{balance}    
\usepackage{booktabs}   
\usepackage{graphics}   
\usepackage{url}        
\usepackage{pifont}     
\usepackage{xcolor}      
\usepackage{xspace}
\usepackage{balance}
\newcommand{\eg}[0]{\textit{e.g.},\ }   

\usepackage[labelfont={bf,small},textfont={bf,small}]{caption}
\usepackage{pifont}

\newcommand{\eat}[1]{}

\usepackage[math]{alp}

\usepackage{algpseudocode}
\usepackage{algorithm}
\usepackage{subcaption}
\usepackage{amsfonts}

\usepackage{amssymb}

\usepackage{amsmath}
\usepackage{color}
\usepackage{enumitem}
\usepackage{graphicx}
\usepackage{epstopdf}
\usepackage{multirow}
\newcommand{\R}{\mathbb{R}}

\usepackage{newfloat}
\usepackage{listings}
\floatstyle{ruled}
\newfloat{listing}{tb}{lst}{}
\floatname{listing}{Listing}

\usepackage{xcolor}
\definecolor{tblue}{RGB}{93, 142, 150}

\begin{document}


\title{SkipSNN: Efficiently Classifying Spike Trains with Event-attention}

\makeatletter
\newcommand{\linebreakand}{%
  \end{@IEEEauthorhalign}
  \hfill\mbox{}\par
  \mbox{}\hfill\begin{@IEEEauthorhalign}
}
\makeatother

\author{
  \IEEEauthorblockN{Hang Yin}
  \IEEEauthorblockA{\textit{Worcester Polytechnic Institute}\\
    Worcester, MA, USA \\
    hyin@wpi.edu}
  \and
  \IEEEauthorblockN{Yao Su}
  \IEEEauthorblockA{\textit{Worcester Polytechnic Institute}\\
    Worcester, MA, USA \\
    ysu6@wpi.edu}
  \and
  \IEEEauthorblockN{Liping Liu}
  \IEEEauthorblockA{\textit{Tufts University}\\
    Medford, MA, USA \\
    liping.liu@tufts.edu}
  \linebreakand 
  \IEEEauthorblockN{Thomas Hartvigsen}
  \IEEEauthorblockA{\textit{University of Virginia}\\
    Charlottesville, VA, USA \\
    hartvigsen@virginia.edu}
  \and
  \IEEEauthorblockN{Xin Dai}
  \IEEEauthorblockA{\textit{Worcester Polytechnic Institute}\\
    Worcester, MA, USA \\
    xdai5@wpi.edu}
  \and
  \IEEEauthorblockN{Xiangnan Kong}
  \IEEEauthorblockA{\textit{Worcester Polytechnic Institute}\\
    Worcester, MA, USA \\
    xkong@wpi.edu}
}

\maketitle

\begin{abstract}
Spike train classification has recently become an important topic in the machine learning community, where each spike train is a binary event sequence with \emph{temporal-sparsity of signals of interest} and \emph{temporal-noise} properties.
A promising model for it should follow the design principle of performing intensive computation only when signals of interest appear. 
So such tasks use mainly Spiking Neural Networks (SNNs) due to their consideration of temporal-sparsity of spike trains. 
However, the basic mechanism of SNNs ignore the temporal-noise issue, which makes them computationally expensive and thus high power consumption for analyzing spike trains on resource-constrained platforms. 
As an event-driven model, an SNN neuron makes a reaction given any input signals, making it difficult to quickly find signals of interest.  
In this paper, we introduce an event-attention mechanism that enables SNNs to dynamically highlight useful signals of the original spike trains. 
To this end, we propose SkipSNN, which extends existing SNN models by learning to mask out noise by skipping membrane potential updates and shortening the effective size of the computational graph. 
This process is analogous to how people choose to open and close their eyes to filter the information they see.
We evaluate SkipSNN on various neuromorphic tasks and demonstrate that it achieves significantly better computational efficiency and classification accuracy than other state-of-the-art SNNs.

\end{abstract}


\section{Introduction}

\begin{figure}[t]
\centering
\begin{minipage}{1\columnwidth}
    \includegraphics[width=1\textwidth]{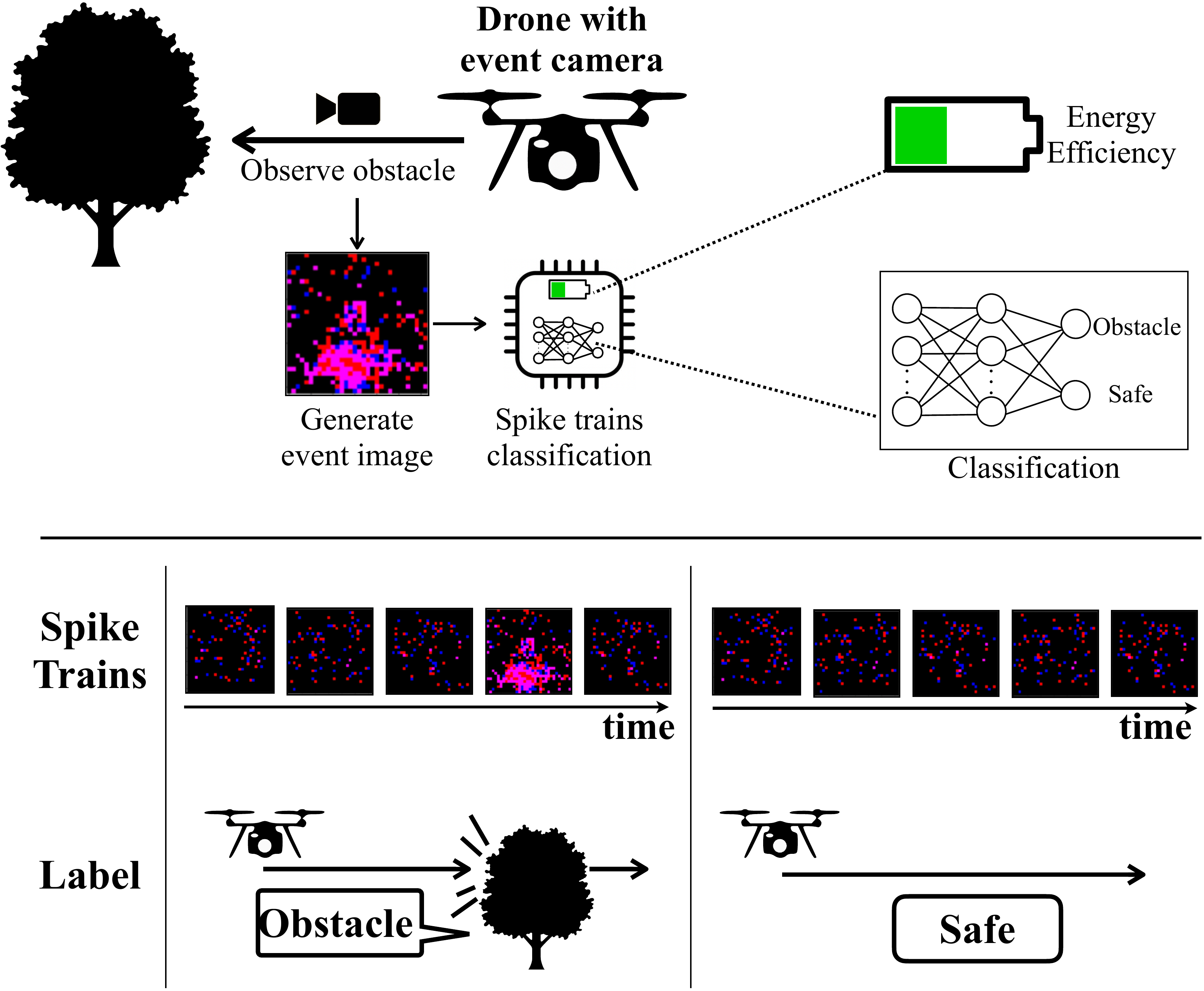}
\end{minipage}
\caption{The problem definition of efficient classification of spike trains. The spike trains are generated by an event camera, which is an imaging sensor that responds to local changes in brightness. Each pixel inside an event camera operates independently and asynchronously, reporting changes in brightness as they occur, and staying silent otherwise. Therefore, each image can be considered as binary event image.}
\label{fig:intro}
\vspace{-12pt}
\end{figure}


\textbf{Motivation.}
Spike trains are sequences of binary signals where 1s are spikes and 0s are not spikes. Such data are common to a variety of domains and are classically analogous to electrochemical signals in the human brain. Spike train datasets are generated from event cameras, which resemble the human eye. Event cameras, also called neuromorphic cameras, require little energy and are designed to capture objects at high speed. Thus, spike train datasets naturally arise during the development of dynamic vision devices \cite{mitra2008real, pfeiffer2018deep, drazen2011toward, mueggler2017event, rebecq2019high}.

Recently, spike train classification has attracted much attention in the machine learning community \cite{wu2019direct, yin2021energy, wu2018spatio,shrestha2018slayer, xin2001supervised, schemmel2006implementing, wade2010swat}. 
Compared with the traditional sequence classification tasks, spike train classification is unique in the following two aspects \cite{gallego2019event, moeys2017sensitive, nozaki2017temperature, brown2004multiple}: 
1) \emph{Temporal-sparsity} of signals of interest. The label of a spike train is only related with certain objects that may only appear in a very small portion of the whole time window.
2) \emph{Temporal-noise} problem. The signals at the majority of time steps are generated from background activities that are not related to objects of interest. 
These two properties of spike train classification impede the application of the widely used deep learning models, \eg recurrent neural networks (RNNs), because of their high computational costs, unnecessarily spent on the whole time window. 
Tasks on spike train data are instead usually processed on energy-sensitive platforms such as wireless monitors and drones, so more computationally efficient models are required. 
To this end, we propose a new \emph{design principle} to guide developement of machine learning models for spike train classification: \emph{perform intensive computation only when signals of interest appear}. 

\textbf{Knowledge Gap.}
Spiking Neural Networks (SNNs) are potential candidates for spike train classification \cite{wu2018spatio,wu2019direct,shrestha2018slayer}, since they are attempt to meet the aforementioned design principle by considering \emph{the temporal-sparsity} of spike trains. 
They take spike trains as inputs and outputs, using biologically inspired, event-driven computation and communication in their design. 
An SNN neuron's core function is to react only when its cummulative membrane potential exceeds a fixed value.
As a result, the neuron has a chance to be activated only when it currently has an event signal, which is passed in as a binary spike. 
Thus, compared to traditional deep learning models, SNNs can build large-scale neural networks with far less energy and memory for spike train classification. 

However, SNNs primarily consider the \emph{temporal-sparsity} of spike trains, overlooking the critical aspect of \emph{temporal-noise}.
As illustrated in Figure \ref{fig:intro}, consider a scenario where a drone equipped with an event camera detects obstacles to adjust its route.
For most of the operational timeline, the camera records signals irrelevant to obstacle detection. Nevertheless, SNNs respond to any detected signals, even if they are merely noise. Consequently, SNNs often fail to adhere to the principle that models should activate only in response to signals of interest. We find that this misalignment is a fundamental factor contributing to the poor generalization and decreased energy efficiency observed in SNNs in real-world applications.

To achieve high classification accuracy with low computational cost for real-world spike trains, we need to follow the aforementioned design principle by considering both temporal-sparsity of useful signals and temporal-noise issue. 
An intuitive approach, which we pioneer in this paper, is for the model to stop processing data when the relevant object is out of its field of view. This behavior is analogous to how we open and close our eyes to filter out the information we see.

\textbf{Challenges.}
We propose a novel method for allowing SNNs to efficiently classify spike trains. Solving this problem is challenging for two main reasons:
\begin{itemize}
    \item \textit{Neuron Consistency}: 
    The promise of SNNs comes from their likeness to
    real neural circuits in the human brain.
    Maintaining this similarity is essential to successful SNNs.
    However, in the standard SNN when a neuron enters a hibernation state, it is hard to wake it up again if there is no new signal input to the network. 
    This means that if the model ignores the input at a timestep, the neurons in the network will lack new input signals. 
    This keeps the neurons silent, making the model likely to ignore potentially useful signals in the future. 
    Thus, when extending SNNs to our problem setting, it is challenging to train successful spiking neurons to skip updates.
    
    \item \textit{Non-differentiability}: SNNs are notoriously difficult to train due to non-differentiable nature of spike activity. 
    Most related works use the rectangular function or sigmoid to approximate the corresponding derivative. 
    However, in practice we find that when our optimization objective considers both accuracy and efficiency, this approximation leads to decayed performance. Additionally, optimization largely depends on the initial values of the parameters of the model.
    Even though some parameter initialization methods such as Glorot \cite{glorot2010understanding} work for traditional artificial neural networks, they lack theoretical basis in SNNs. 
    Designing an efficient optimization algorithm is the second challenge we face.
\end{itemize}

\begin{figure}[t]
  \centering
  \includegraphics[width=\linewidth]{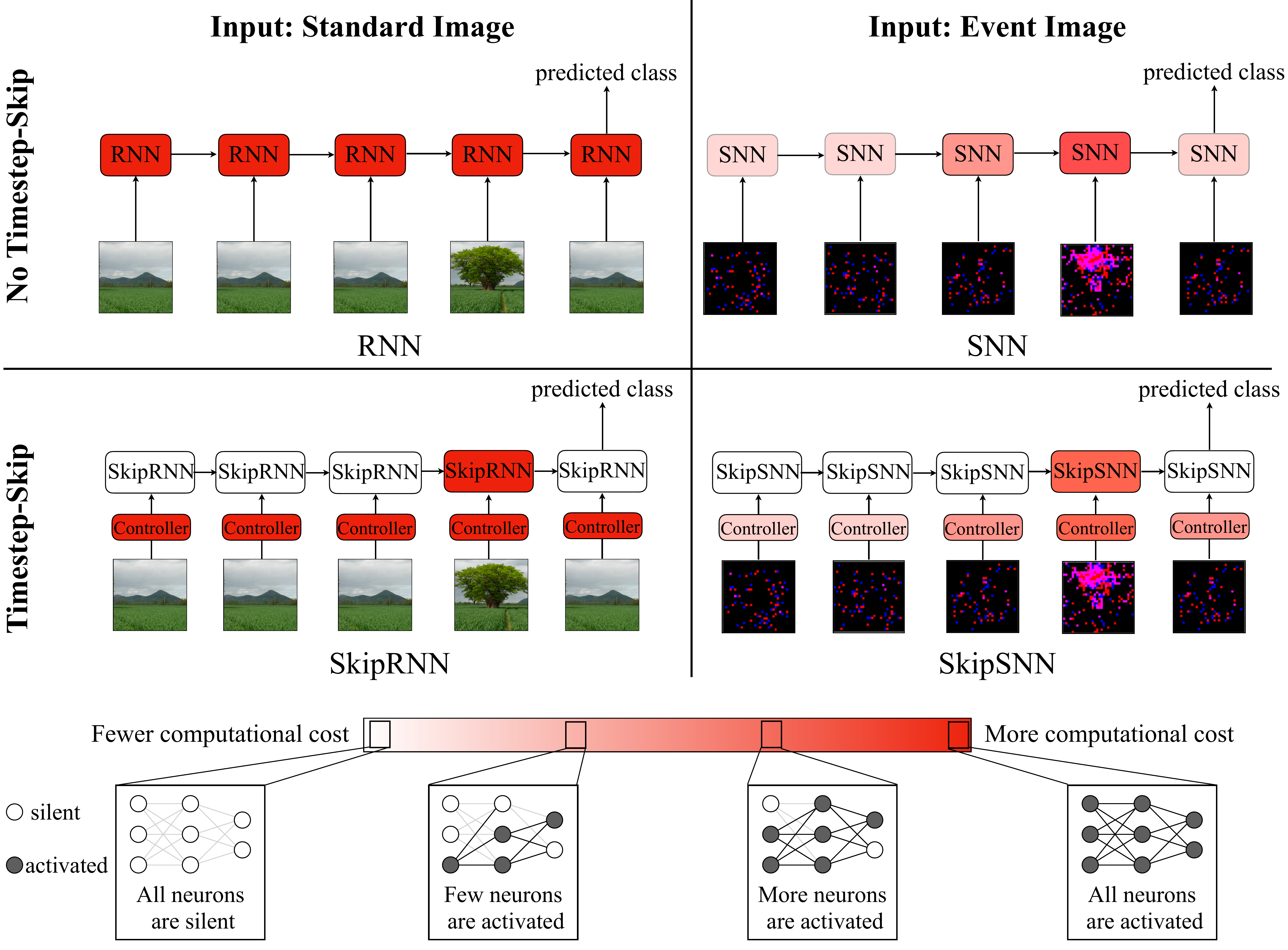}
  \vspace{-10pt}
  \caption{Differences among Recurrent Neural Network (RNN) , SkipRNN \cite{campos2017skip}, Spiking Neural Network (SNN) \cite{wu2018spatio, shrestha2018slayer} and SkipSNN (ours). SkipSNN outperforms others in computational efficiency and classification accuracy with an event-attention mechanism for noise filtering.}
  \label{fig:intro2}
  \vspace{-12pt}
\end{figure}

\textbf{Proposed Method.}
To solve our problem,
we introduce an event attention mechanism that enables SNNs to dynamically highlight useful signals in the input spike train. 
We extend existing SNNs to have two different states: \textit{awake} and \textit{hibernating}, inspired by how people's eyes open and close, turning on and off data intake. If our SNN enters its awake state at time step $t$, it will consider the input at $t$. Otherwise, if it hibernates at time step $t$, it will ignore the input at $t$. 
To this end, we design a controller that switches the model between these two states. 
Since this is not differentiable, we also introduce a new loss function with a penalty that trades off accuracy and computational cost.
In this way, our extended SNN learns to mask out noise by skipping updates and shorten the effective size of the computational graph without requiring any additional supervision signal. 
We refer to our model as SkipSNN and illustrate the difference between it and traditional SNN in Figure \ref{fig:intro2}.

\textbf{Contributions.} 
Our key contributions are as follows:
\begin{itemize}
    \item We define the problem and modeling principle of general spike train classification, which is important for smart dynamic sensor systems with limited energy.
    \item We propose SkipSNN, which solves this problem and can be used on energy-limited dynamic sensor devices. \item We develop an efficient optimization technique to train our SkipSNN model.
    \item We demonstrate that our model outperforms recent state-of-the-art alternatives by achieving higher accuracy and lower computational cost when tested on both the neuromorphic MNIST and DVS-Gesture datasets. 
\end{itemize}

The rest of our paper is organized as follows. First, we review related work, then introduce details of the background methods. Next, in Section 4, we present our proposed method. We then describe our experimental setup and discuss our results in Section 5. Finally, we conclude the paper with key take-aways and give some directions for future work.

\section{Related Work}
In spike train classification, methods for training SNNs can be divided into two categories: ``Indirect" learning and ``Direct" learning. 
Indirect learning mainly focuses on ANN-to-SNN conversion \cite{diehl2015fast,cao2015spiking,hu2018spiking, midya2019artificial}.
These methods are indirect in that a regular non-spiking Artificial Neural Network (ANN), such as a multi-layer perceptron, is initially used during the training phase. 
At inference-time, the trained model is then converted to an SNN. However, such indirect training doesn't align well with how an SNN operates. For example, in ANNs, it does not matter if activations are negative, while firing rates in SNNs are always positive. As for Direct learning, many methods have recently been proposed \cite{shrestha2018slayer, wu2019direct, yin2021energy}. These approaches train SNNs directly using back-propagation in both the spatial and temporal domains. \cite{shrestha2018slayer,wu2019direct} have achieved state-of-the-art accuracy on the MNIST and N-MNIST datasets. 
\cite{yin2021energy} trains SNNs with a spatial sparsification technique that allows SNNs to perform inference with lower computational cost. While these methods perform better than the indirect methods on many neuromorphic datasets, they are still not suitable for efficient classification of real-world spike trains, especially in terms of temporal-noise problem. 

Although there is no model for dealing with our defined problem for SNNs, some methods have been proposed for dealing with similar problems for RNNs \cite{lalapura2021recurrent, lei2021attention, morais2021learning, zheng2021accurate, hartvigsen2020learning, jernite2016variable, campos2017skip, shen2018ordered}. 
For example, SkipRNN \cite{campos2017skip} extends classic RNN models by learning an additional controller network that learns to skip state updates. 
The input of this neuron is the state value of other neurons. So it can generate a binary value based on the sigmoid function value of its state. This model can significantly reduce the computational cost. 
However, while both RNNs and SNNs are used to deal with sequence analysis, they remain completely different. 
The main difference is that neurons in RNNs are mostly non-linear, continuous function approximators that operate on a common clock cycle, whereas the neurons in SNNs use asynchronous spikes that signal the occurrence of some characteristic event and temporally precise membrane potentials. When dealing with spiking training tasks, many neurons of an SNN may stay silent based on its mechanism, whereas all neurons of an RNN will be activated. 
Consequently, in terms of inference efficiency, RNNs are outmatched by SNNs for spike train classification. 
We illustrate this key difference in Figure~\ref{fig:intro2}.

\begin{figure*}[t]
\centering
\begin{minipage}{2\columnwidth}
    \includegraphics[width=0.95\textwidth]{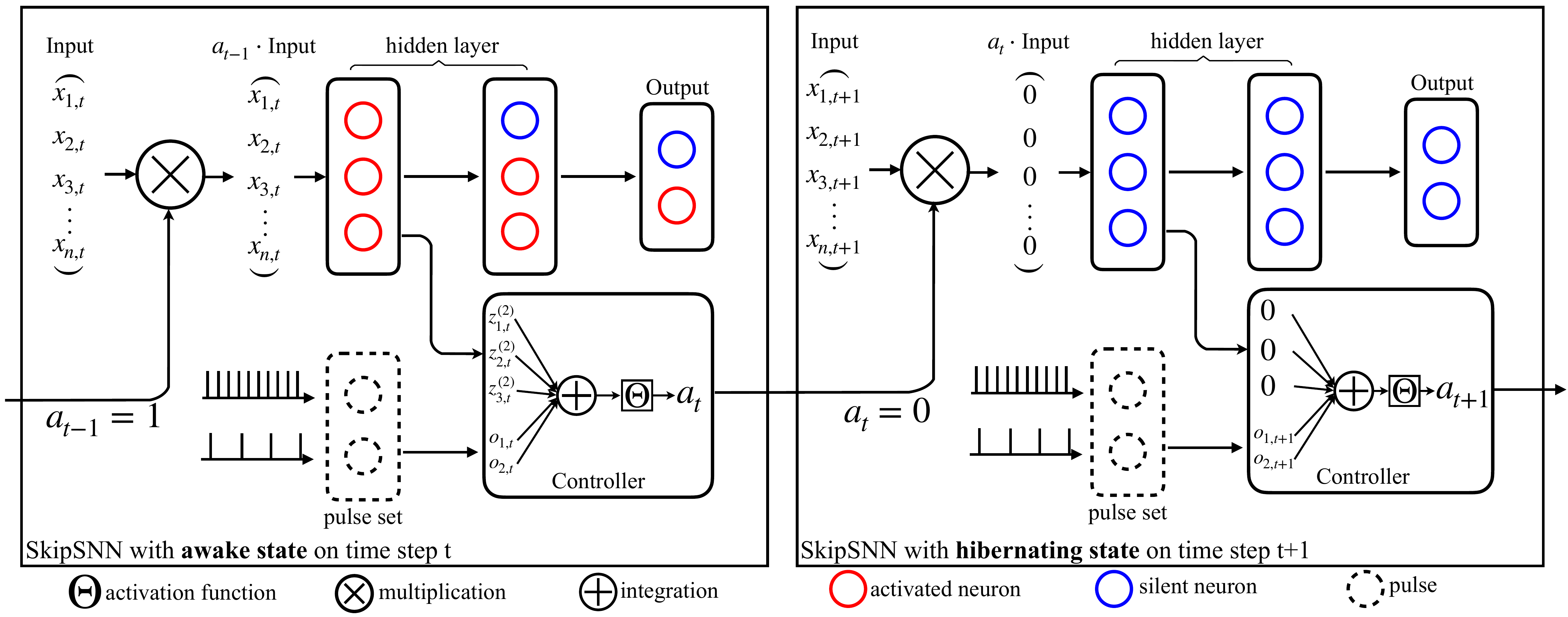}
\end{minipage}
\caption{Model architecture of the proposed SkipSNN.}
\label{fig:model}
\vspace{-15pt}
\end{figure*}
\section{PRELIMINARIES} 
We begin by describing a one-neuron SNN that accepts a one-feature input. 
This neuron is a recurrent unit that is affected by current input, and previous output and membrane potential of itself. For each timestep $t$, it takes $x_t$ as input and combines its previous membrane potential $u_{t-1}$ and output $z_{t-1}$ to update $u_t$. For the activation of spiking neuron, 
when $u_t$ exceeds threshold $V_{\text{th}}$, the neuron will fire at timestep $t$.
We get the output $z_t$ by the step function $\mathrm{\Theta(\cdot)}=0$, which satisfies $\mathrm{\Theta(x)}=0$ when $x < 0$, otherwise $\mathrm{\Theta}(x)=1$. Therefore, for each timestep $t$, the membrane potential $u_t$ and output $z_t$ are expressed as follows:
\begin{align}
\label{simple1}
    &u_{t} = \tau u_{t-1}(1-z_{t-1})+w x_{t},\\
    \label{simple2}
    & z_{t} = \mathrm{\Theta}(u_t-V_{\text{th}}),
\end{align}
where hyperparameter $\tau \in [0,1]$ denotes a time decay constant and $w$ is the connection between input and this neuron. 

The SNN expressed in (\ref{simple1}-\ref{simple2}) more closely resembles natural neural networks than traditional ANNs do.
In this way, we represent neuron as a parallel combination of a ``leaky'' resistor and a capacitor. The second term of the r.h.s. of (\ref{simple1}) is used as external current input to charge up the capacitor to update membrane potential $u_t$. If the neuron emits a spike $z_t=1$ at timestep $t$, the capacitor discharges to a resting potential (which we fix at zero throughout this paper) by using the first term in (\ref{simple1}).

A neural network is built by hooking together many of these simple ``neurons'' so that the output of a neuron is the input of another. 
We let $n_\ell$ denote the number of layers in our network and label layer $l$ as $L_l$, so the input layer is $L_1$, and the output layer is $L_{n_\ell}$. Our network has parameter set $\bW = \{\bW^{(1)},\dots,\bW^{(n_\ell-1)},\}$, where $\bW^{(\ell)}_{ij}$ denotes the parameter associated with the connection between neuron $j$ in layer $\ell$, and neuron $i$ in layer $\ell+1$. We also let $s_\ell$ denote the number of neurons in layer $\ell$. For layer $\ell \in \{2,\dots,n_\ell\}$, We write $\bu_{t}^{(\ell)}=(u_{i,t}^{(1)},\ldots,u_{i,t}^{(s_\ell)})^\top$ and $\bz_{t}^{(\ell)} = (z_{i,t}^{(1)},\ldots,z_{i,t}^{(s_\ell)})^\top$ to denote the state and output vector of neurons at timestep $t$. For $\ell=1$, we will use $\bz^{(1)}_t=\bx_t$ to denote the input vector. Given a fixed setting of the parameter $\bW$ and the threshold $V_{\text{th}}$ of each neuron, the update of neurons in all layers is given by:
\begin{align}
\label{vector}
    & \bu^{(\ell)}_t = \tau \bu_{t-1}^{(\ell)} \odot (1 - \bz_{t-1}^{(\ell)}) + \bW^{(\ell-1)} \bz^{(\ell-1)}_t,\\
    \label{vector2}
    & \bz^{(\ell)}_t = \mathrm{\Theta}(\bu^{(\ell)}_t-V_{\text{th}}).
\end{align}

Given the expressions above, we can solve a standard SNN classification problem by training a classifier $f$: ${\R}^{P\times T} \mapsto \{1,\ldots,N\}$ on a given dataset $\{(\bx^{(1)},y^{(1)}),\dots,(\bx^{(K)},y^{(K)})\}$ that contains $K$ training samples, of which each instance $\bx^{(i)} \in {\R}^{P\times T}$ has an observed label $y^{(i)} \in \{1,\ldots,l(N)\}$. $P$ is the number of input entries and $T$ denotes the length of spike train. To train such SNN, we can minimize the following loss function $L$ for a single training example $(\bx,y)$:
\begin{align}
\label{eq:mse}
L = \left(y - \frac{1}{T}\sum^{T}_{t}\mathbf{M}\bz^{(n_{\ell})}_t\right)^{2}
\end{align}
where $\bz^{t,N}$ denotes the voting vector of the last layer $N$ at time step $t$, $\mathbf{M}$ denotes a constant voting vector connecting neurons in the output layer to a specific class.

From Equations \ref{vector}-\ref{vector2}, spike signals propagate not only through the layer-by-layer spatial domain but also influence neuronal membrane potentials through the temporal domain. Thus, both spatial and temporal directions are considered during error backpropagation, \textit{i.e.}, spatio-temporal backpropagation (STBP) \cite{wu2018spatio,wu2019direct}, which significantly enhances network accuracy. During backpropagation, since the activity function $\mathrm{\Theta}(\cdot)$ is non-differentiable, the rectangular function is commonly used to approximate its derivative.


\section{Methodology}

In this paper, we propose a novel modification for existing SNN architectures that allow them to mask noise and skip membrane potential updates without requiring any additional supervision signal. 
To build a new architecture, which we call SkipSNN, we view SNN neurons as having two different states: awake and hibernating. Then, we design a new neuron to control the model to switch freely between these two states. 

Inspired by \cite{campos2017skip}, we augment the network with a binary gate,
but use a novel neuron to control it.
This neuron doesn't have any special settings, but follows the basic mechanism of SNNs, exactly like other neurons in the network. 
In the rest of this paper, in order to distinguish it from other SNN neurons, we call this neuron the \textit{controller}.

\textbf{Model Description.}
Let $v_t$ denote the membrane potential of the controller at time step $t$. 
This controller is connected with all the neurons in the first layer. 
It is therefore affected by the outputs of the first layer and will generate a binary output $a_t$ according to the SNN mechanism described in Equations \ref{simple1} and \ref{simple2}. 
Then, this value decides whether to consider the input of the next time step by treating $a_t$ as a multiplier of the next time step. 
If it emits a spike, this means that the network enters the awake state at next time step. 
Otherwise, the network hibernates at next time step. 
Based on the SNN mechanism, the controller will reset its membrane potential to zero after emit a spike. 
Therefore, when it decides to enter the awake state at the next time step, it is hard for the controller to spike again due to the lack of new spike inputs from the first layer.

In order not to miss the potential useful signals in the future, a temporal form of bias is needed to adjust the membrane potential of the controller. 
In this paper, we introduce synchronisation pulses that act as additional inputs to the controller, in order to provide the information of time.
These can be thought of as similar to internally-generated rhythmic activity in biological networks, such as alpha waves in the visual cortex \cite{klimesch2012alpha} or theta and gamma waves in the hippocampus \cite{buzsaki2006rhythms}.

In our proposed SkipSNN, a set of pulses are fully connected to the controller in the network. Each pulse spikes at a different frequency.
For example, some spike once every time step, some spike once every 10 time steps, and some spike once every 100 time steps. 
Subsequently, these pulses can modify the membrane potential of the controller when it stays hibernating state, thereby being activated again. 
At every time step $t$, affected by the output from the neurons in the first hidden layer and pulses, the controller emits a binary signal, which is multiplied by the model input at $t+1$. 
The resulting architecture depicted in Figure \ref{fig:model} can thus be described as follows:
\begin{align}
\label{control}
    & \bu^{(2)}_t = \tau \bu_{t-1}^{(2)} \odot (1 - \bz_{t-1}^{(2)}) + a_{t-1}\bW^{(1)} \bx_t,\\
    \label{control2}
    & v_t = \tau v_{t-1} (1 - a_{t-1}) + \bW_{\text{z}} \bz_t^{(2)} + \bW_{\text{o}} \bo_t,\\
    \label{control3}
    & a_t = \mathrm{\Theta}(v_t-V_{\text{th}}),
\end{align}
where $\bW_{\text{z}}$ and $\bW_{\text{o}}$ are the weights vectors between the controller and neurons in the first hidden layer, and the controller and pulses, respectively. 
$\bo_t = (o_{1,t}, ..., o_{p,t})$ is the pulse vector that contains the signals from $p$ different pulses at t. 
$v_t$ and $a_t$ are the membrane potential and the output of the controller at t, respectively. 

According to the model formulation from Equations \ref{control}-\ref{control3}, SkipSNN can switch between awake and hibernating state based on the value of $a_t$. 
If $a_t = 1$, SkipSNN will be updated based on the input $\bx_{t+1}$ at $t+1$, otherwise, the proposed model will skip $\bx_{t+1}$ due to $a_t \bx_{t+1} = 0$.

In particular, the controller can be connected to any layer of SNN. 
Meanwhile, the controller itself can also be a multilayer SNN. 
However, in practice, we found that the controller works better when connected to the first hidden layer. 
In addition, using a multi-layer network structure as the controller does not improve the performance but will increase the computational cost.
Thereby, in this paper, we keep using the network structure depicted in Figure \ref{fig:model}.

\textbf{Limiting Computation.}
The proposed SkipSNN learns when to enter awake or hibernating state without requiring any additional supervision signal. 
The longer the model stays hibernating state, the lower computational cost required during inference phase. 
However, there is a trade-off between classification accuracy and computational power. 
If the model sacrifices some important time steps that contain useful information, it can reduce the computational cost, but it will also sacrifice the accuracy of the model. 
To balance between accuracy and efficiency, we add an additional penalty term
\begin{align}
\label{penalty}
    L_{\text{penalty}} = \lambda \frac{\sum^T_{t=1} a_t}{T},
\end{align}
where $L_{\text{penalty}}$ is the cost associated to one single sample, $\lambda$ is the cost per sample, and $T$ is the spike train length.

\textbf{Optimizing SkipSNN.}
SNN is hard to optimize because the derivative of its activation function is a $\delta$ function, whose value is zero everywhere except at threshold. 
As is done in prior works \cite{wu2018spatio,wu2019direct}, we could use the rectangular function to approximate the corresponding derivative. 
However, the switch between awake and hibernating state is directly determined by the output of the controller. 
When the corresponding gradient of its membrane potential is zero, we will not be able to continue to optimize the relevant parameters of the controller through gradient descent. 
Therefore, we propose simulated annealing for SkipSNN; the training process for SkipSNN is divided into two stages.

In the first stage, we set $\lambda = 0$, which will cause the model to stay in the awake state at all times.
This means that the controller will always emit $a_t=1$ during this stage, because its membrane potential $v_t$ is always larger than the threshold $V_{\text{th}}$. 
Therefore, our goal at this stage is to make SkipSNN solve the classification task as accurately as possible. 
In this stage, the derivative of each activation is approximated by the rectangular function denoted by $h(u)$:
\begin{align}
\label{approx1}
    h(u) =  \frac{1}{\epsilon}\text{sign}\left(|u-V_{\text{th}}|<\frac{\epsilon}{2}\right),
\end{align}
where $\epsilon$ determines the peak width.

\begin{algorithm}[t]
\caption{Algorithm for \texttt{SkipSNN}}
\label{alg_skip}
\begin{algorithmic}[1]
\Require 

i: Network inputs $\{X^{t}\}^{T}_{t}$;

ii: class label $Y$;

iii: parameters and states of main network $(\{\bW^{(\ell)},\bu^{(\ell)}_0,\bz^{(\ell)}_0\}^{N_{1}-1}_{\ell=1})$;

iv: parameters and states of controller
$(\bW_z,\bW_o,\bv_t,\ba_0)$;

v: the parameters of iterative LIF and penalty ($T, \epsilon, \Delta, V_{th}, \lambda$);

vi: $\text{iter}_{max}$: the maximum number of iteration

\Ensure: Update network parameters

\hspace{-25pt}\textbf{Stage 1}:
        \begin{algorithmic}[1]
        \State Set the approximation of activation with Eq.\ref{approx1}
        \State Set $\lambda = 0$
        \Repeat
        \State Update the parameters of main network $(\{\bW^{(\ell)}\}^{N_{1}-1}_{\ell=1})$ 
        \Until{$iter={iter}_{max} \text{ or convergence}$}
        \end{algorithmic}

\hspace{-25pt}\textbf{Stage 2}:
        \begin{algorithmic}[1]
        \State freeze $(\{\bW^{(\ell)}\}^{N_{1}-1}_{\ell=1})$ 
        \State Set the approximation of activation with Eq.\ref{approx2}
        \State Set $\lambda > 0$
        \Repeat
        \State Update the parameters of controller $(\bW_z,\bW_o)$
        \Until{$iter={iter}_{max} \text{ or convergence}$}
        \end{algorithmic}
\end{algorithmic}
\end{algorithm}
In the second stage, we freeze parameters that are not related to the controller, and start optimizing the controller by elevating the multiplier $\lambda$ of the penalty loss. 
Because $v_t$ at that point is always larger than $V_{\text{th}}$, we deploy a sigmoid function instead of rectangular function to approximate its derivative:
\begin{align}
\label{approx2}
    h(u) =  \frac{1}{1 + e^{\frac{1}{\Delta}(u-V_{\text{th}})}}.
\end{align}
The parameter $\Delta$ in Equation \ref{approx2} controls the steepness of the sigmoid function, which is considered to be the reciprocal pseudo-temperature. 
As $\Delta \to 0$, the sigmoid becomes a step function and the stochastic unit becomes deterministic. As $\Delta$ increases, this sharp threshold is ``softened'', thus making the range of the sigmoid wider. 
Therefore in this stage we initialize $\Delta$ with a high value to make sure the controller has a gradient when the membrane potential is high. 
Then, we decrease $\Delta$ periodically during the optimization process of the parameters related to the controller. 
We also summarize the overall training process of our proposed SkipSNNs
as pseudo-code in Algorithm \ref{alg_skip}.

\section{Empirical Study}
To comprehensively validate the effectiveness of our proposed method, we conduct experiments to answer two research questions: First, we are interested in accuracy improvement on classification problem of spike trains; Second, we want to demonstrate that our model can significantly reduce computational cost with very negligible degradation in accuracy. 
As ours is the first SNN proposed to deal with our defined problem, we compare our model with Fixed-skip SNN and Random-skip SNN. 
We also use a modified SNN converted from SkipRNN \cite{campos2017skip} as a baseline, because SkipRNN is a cutting-edge model for skipping frames on time series, which is similar to our defined problem.
To better compare our work with them, we test on various neuromorphic datasets, including N-MNIST and DVS-Gesture. Both are widely used to evaluate SNN models in related work. 
To combat randomness in the experiment system, we run all experiments 10 times and report the average results, except when otherwise stated. The source code is publicly available at~\url{https://github.com/Anonymous6369/SkipSNN}.

\begin{figure}[t]
\centering
\vspace{-5pt}
\subfigure[Accuracy on N-MNIST]{
  \includegraphics[width=0.525\columnwidth]{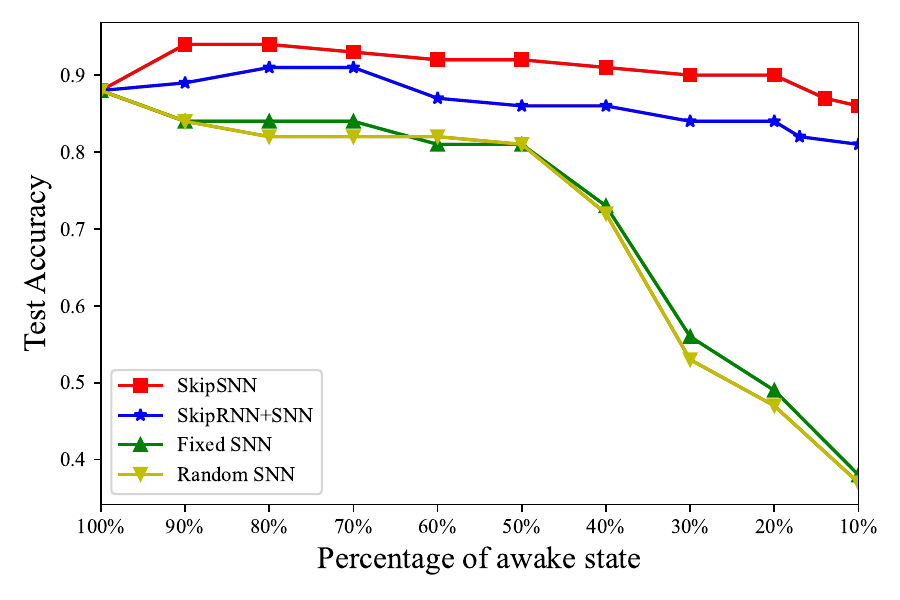}
}
\subfigure[Accuracy on DVS-Gesture]{
  \includegraphics[width=0.525\columnwidth]{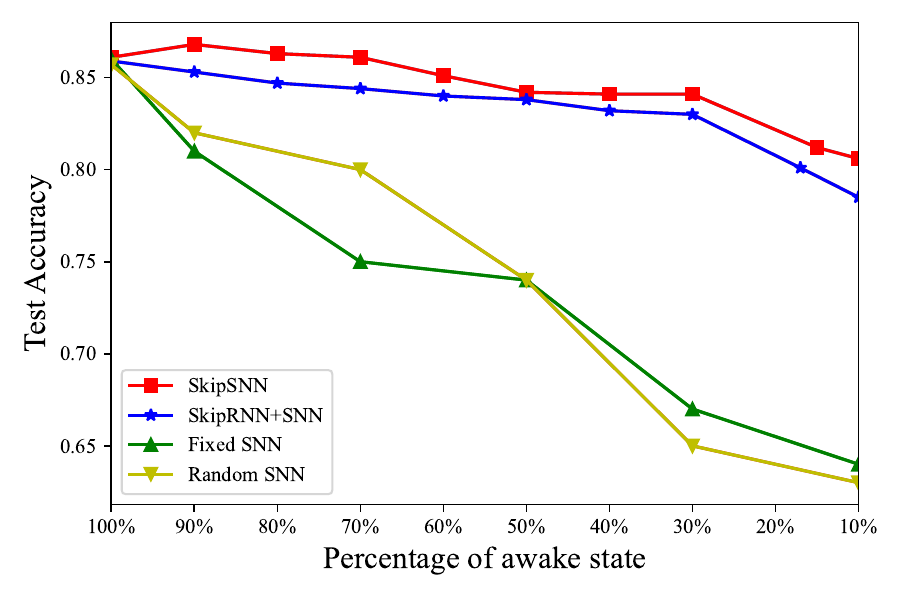}
}
\caption{Observing the performance of different models with different percentages of updated time-steps.}
\label{fig:lambda}
\vspace{-12pt}
\end{figure}

\subsection{Datasets}
We evaluate our proposed model and baselines on various datasets. To simulate the properties of general spike trains in the real world, we modified these neuromorphic datasets (N-MNIST\footnote{https://www.garrickorchard.com/datasets/n-mnist} and DVS-Gesture\footnote{https://ibm.ent.box.com/s/3hiq58ww1pbbjrinh367ykfdf60xsfm8}) by making signals of interest temporal-sparse and adding noise into them. 

\subsubsection{N-MNIST}
The N-MNIST dataset \cite{orchard2015converting} consists of MNIST images converted into a spiking dataset using a Dynamic Vision Sensor (DVS) moving on a pan-tilt unit. In our experiments, each dataset sample is $50$ ms long, with a shape of $34 \times 34$ pixels, containing two channels to preserve ``on'' and ``off'' spikes, respectively. 
This dataset is harder than MNIST because one has to deal with saccadic motion.
For SkipSNN experiments, 
we generate a $300$ ms blank time sequence firstly, meanwhile increasing the noise by adding one signal to a random pixel of the image in each millisecond (one time step). 
Then we put each N-MNIST sample into a random period of each time sequence. 
So in each final sequence, only $16.7$\% of the time steps have useful signals related to original N-MNIST dataset.
The dataset is split into $60,000$ training samples and $10,000$ testing samples. 

\subsubsection{DVS-Gesture}
The DVS-Gesture dataset \cite{amir2017low} contains $1,342$ instances of a set of $11$ hand and arm gestures, grouped into $122$ trials and collected from $29$ subjects under $3$ different lighting conditions. 
During each trial, one subject stood against a stationary background and performed all $11$ gestures sequentially under the same lighting conditions. 
The problem is to identify the correct action label associated with each action sequence video. 
In our experiments, each DVS-Gesture sample is $400$ ms long, and $32 \times 32$ pixels big, containing two channels to preserve ``on'' and ``off'' spikes.
To allow us to test SkipSNN on spike trains, we modify DVS-Gesture dataset by using the method applied to modify N-MNIST. 
So each spike train in the final dataset has 1000 time steps (ms), and only 400 consecutive time steps in each spike train have useful signals related to original DVS-Gesture dataset.

\begin{table*}[t!]
\centering
\caption{Comparative results on N-MNIST dataset.}
\vspace{-2pt}
\label{tab:nmnist}
\begin{tabular}{cccc}
\hline
{Model} & {Percentage of awake state} & {Accuracy}&{Inference MFLOPs}\\
\hline
{SNN}    & 100\% $\pm$ 0.0\% & 88.92\% $\pm$ 0.76\% & 1.15 \\ {Fixed-skip SNN}    & 13.33\% $\pm$ 0.0\% & 38.43\% $\pm$ 4.13\% & 0.51 \\
{Random-skip SNN}    & 10.00\% $\pm$ 0.0\% & 37.16\% $\pm$ 3.41\% & 0.47 \\
{SkipRNN+SNN, $\lambda=10^{-3}$}    & 91.72\% $\pm$ 0.37\% & 89.86\% $\pm$ 0.16\% & 0.71 \\
{SkipRNN+SNN, $\lambda=10^{-1}$}    & 11.13\% $\pm$ 1.24\% & 81.24\% $\pm$ 1.02\% & 0.52 \\
\hline
{SkipSNN, $\lambda=10^{-3}$}    & 92.15\% $\pm$ 0.08\% & 94.47\% $\pm$ 0.12\% & 0.75 \\
{SkipSNN, $\lambda=10^{-1}$}    & 11.03\% $\pm$ 0.58\% & 86.65\% $\pm$ 0.27\% & 0.46 \\
\hline
\end{tabular}
\end{table*}
\begin{table*}[t!]
\centering
\caption{Comparative results on DVS-Gesture dataset.}
\vspace{-2pt}
\label{tab:dvs}
\begin{tabular}{cccc}
\hline
{Model} & {Percentage of awake state} & {Accuracy}&{Inference MFLOPs}\\
\hline
{SNN}    & 100\% $\pm$ 0.0\% & 86.12\% $\pm$ 0.25\% & 1232.5 \\ {Fixed-skip SNN}    & 13.33\% $\pm$ 0.0\% & 64.13\% $\pm$ 2.05\% & 147.4 \\
{Random-skip SNN}    & 10.00\% $\pm$ 0.0\% & 63.41\% $\pm$ 2.37\% & 112.5 \\
{SkipRNN+SNN, $\lambda=10^{-5}$}    & 92.41\% $\pm$ 0.42\% & 85.32\% $\pm$ 0.43\% & 1013.1 \\
{SkipRNN+SNN, $\lambda=10^{-4}$}    & 10.02\% $\pm$ 3.35\% & 78.50\% $\pm$ 0.43\% & 117.3 \\
\hline
{SkipSNN, $\lambda=10^{-6}$}    & 89.63\% $\pm$ 0.23\% & 86.82\% $\pm$ 0.13\% & 973.7 \\
{SkipSNN, $\lambda=10^{-4}$}    & 9.04\% $\pm$ 1.44\% & 80.24\% $\pm$ 0.18\% & 72.6 \\
\hline
\end{tabular}
\vspace{-12pt}
\end{table*}
\begin{figure}[t]
\centering
\vspace{-5pt}
\subfigure[Examples of N-MNIST]{
  \includegraphics[width=1\columnwidth]{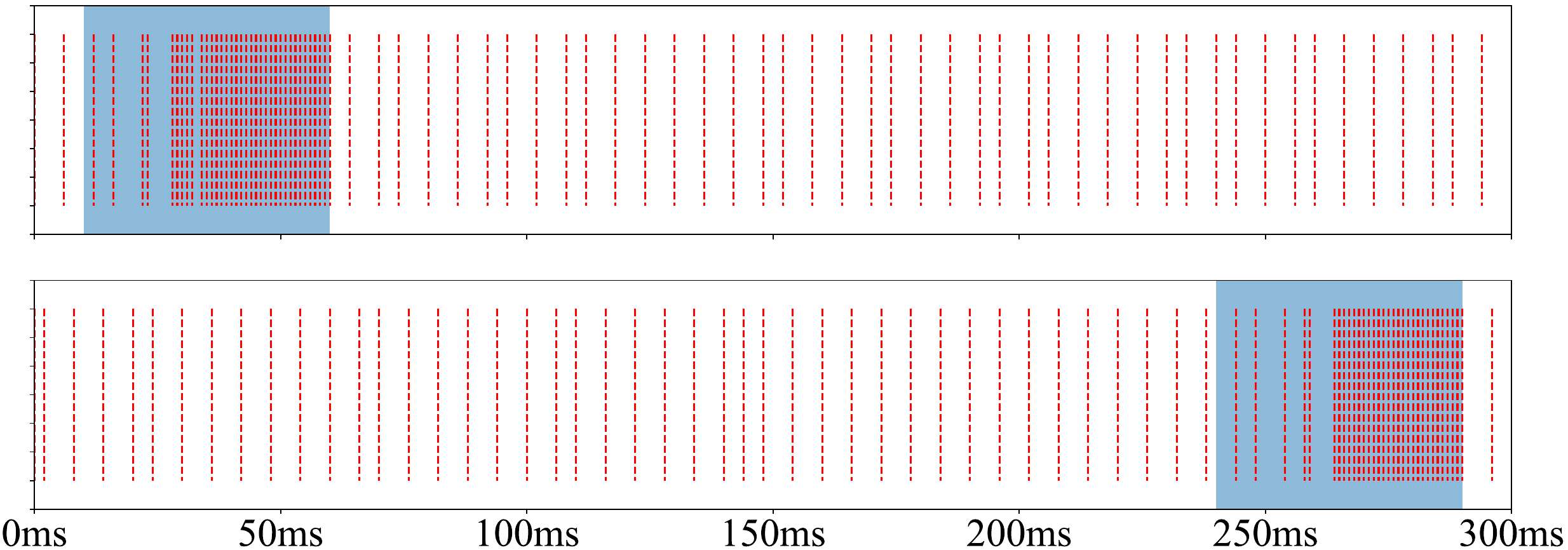}
}
\subfigure[Examples of DVS-Gesture]{
  \includegraphics[width=1\columnwidth]{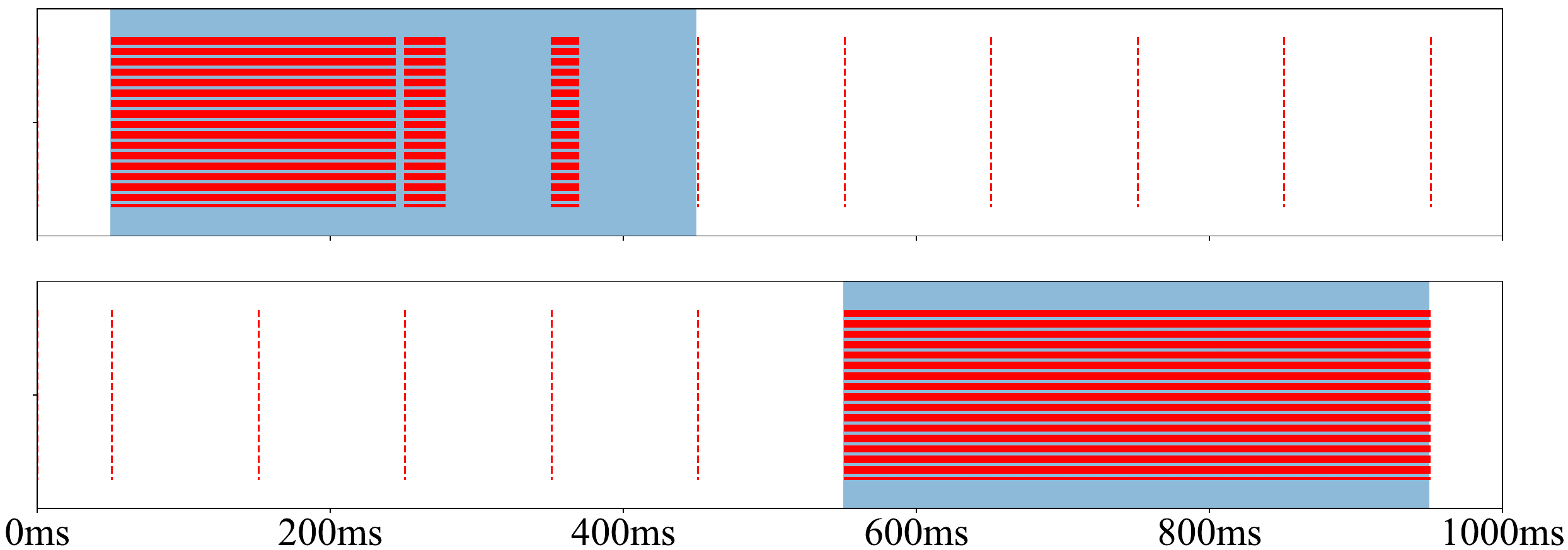}
}
\caption{Temporal raster plot of spikes in the controller neuron during the inference of the examples in N-MNIST and DVS-Gesture by SkipSNN. The spikes shown are generated from the controller, and used to control when to enter awake state. The blue box represents the period of useful signals. The dash red line represents the location of awake state.}
\label{fig:raster}
\vspace{-12pt}
\end{figure}
\subsection{Compared Methods}
To demonstrate the effectiveness of SkipSNN, we test against several baselines:
\begin{itemize}
    \item \textit{Fixed-skip SNN}: Control the percentage of awake state according to the fixed time-step size. For example, $90\%$ means skip $1$ time step after every $9$ time step updates; $50\%$ means skip one time step after every one time step.
    \item \textit{Random-skip SNN}: Determine awake and hibernating state through Bernoulli sampling. We control the percentage of awake states by changing the parameter of Bernoulli distribution.
    \item \textit{SkipRNN + SNN}: SkipRNN \cite{campos2017skip} is proposed to extend existing RNN models by learning to skip state updates and reduce the computational cost. 
    The input and output of SkipRNN are not spike trains. But due to the similarity of the defined problem, we convert SkipRNN to an SNN version as a competitor. 
\end{itemize}
To better compare our work with them, we use the same network structure and optimization algorithm on each model. 
 
\subsection{Evaluation metrics}
To evaluate classification performance, we use the standard Accuracy metric. 
To evaluate the computational efficiency, we use the million floating point operations (MFLOPs) to measure the potential speedup. MFLOPs are computed by assuming one flop for multiplication and one flop for addition.
In our experiments, we use MFLOPs of each model for one same sample in the inference phase as the evaluation metric.

\subsection{Experiment results}
In this section, we separately discuss the experimental results related to each of the two research questions.

\subsubsection{Better accuracy}
The results shown in Figure \ref{fig:lambda} compare the performance of each model with different percentage of updated time-steps on two different datasets. 
We control the percentage of awake state of SkipRNN+SNN and SkipSNN by changing the multiplier $\lambda$ of the time-budget penalty. 
For random-skip SNN, we control it through the parameter of Bernoulli distribution. For fixed-skip SNN, we control it according to different time-step sizes. 
According to the results, we  can  observe that as the percentage of awake state decreases, fixed-skip and random-skip SNN drop rapidly. 
By contrast, SkipRNN+SNN and SkipSNN show their advantages in this scenario. 
Especially, when each model only considers less than $20\%$ of awake state, SkipRNN+SNN and SkipSNN still maintain an accuracy of more than $80\%$. In contrast, the other two models are no longer valid.
According to the comparison of SkipSNN+SNN and SkipRNN, our proposed model has a better performance of classification than the other one on the testing set of both two datasets. 
This robustness in performance directly leads to a significant drop of the computational cost during inference phase. 
Moreover, the accuracy of SkipSNN with the percentage between $90\%$ and $70\%$ is even higher than that without skipping time steps. It means that compared with traditional SNNs, our proposed model is more accurate in dealing with the classification of spike trains. 
\begin{figure}[t]
\centering

\subfigure[Original and reconstructed input of N-MNIST]{
  \includegraphics[width=1\columnwidth]{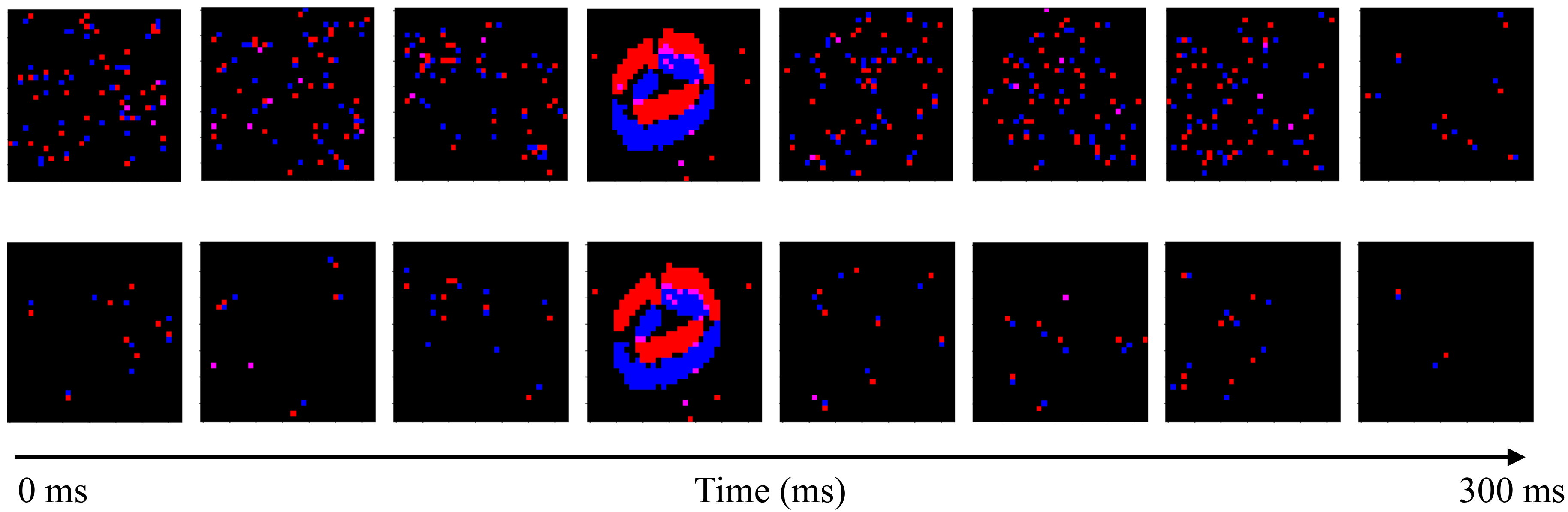}
}
\subfigure[Original and reconstructed input of DVS-Gesture]{
  \includegraphics[width=1\columnwidth]{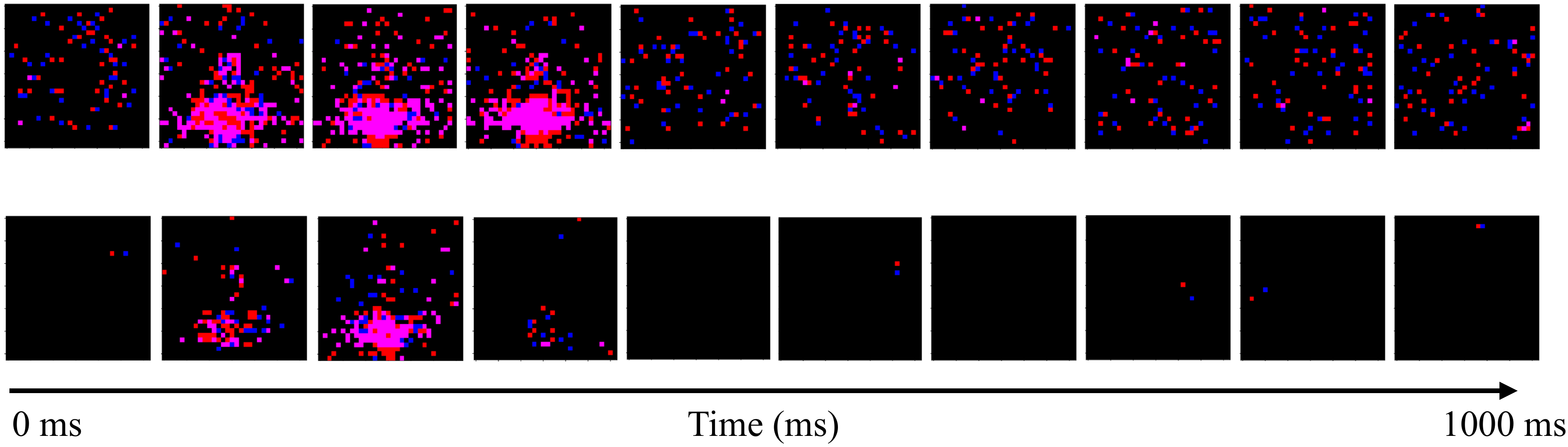}
}
\caption{Visualization of original and reconstructed input of N-MNIST and DVS-Gesture. 
The inputs are, in row order, original and reconstructed. 
Reconstructed input means that we mask the signal in original input according to the time step locations that the model decides to skip.}
\label{fig:recon}
\vspace{-13pt}
\end{figure}

Examples in Figure \ref{fig:raster} illustrate how SkipSNN learns to skip time steps that contain primarily noise or unhelpful for classification. Leveraging the control SkipSNN has over time-step skipping, we visualize its decisions on each time step through a temporal raster plot of spikes in the controller during inference on two datasets. The blue areas in each plot indicate periods of useful signals, while the white areas represent mostly noise. 
Figure \ref{fig:raster} shows that spikes are concentrated in the blue areas and sparse in the white, demonstrating SkipSNN’s ability to distinguish between signals of interest and noise, effectively switching between awake and hibernating states. 

Finally, the reason for the success of our proposed model is that SkipSNN filters out noise with the presence of controller architecture, reconstructing input that is much purer than the original input. 
We demonstrate it in Figure \ref{fig:recon}. 
We show the gif picture of original data and that of reconstructed data, which mask out most of the noise through SkipSNN.

\subsubsection{Potential speedup}

The tables shown in Table \ref{tab:nmnist} and \ref{tab:dvs} compare SkipSNN with a traditional SNN, fixed-skip SNN, random-skip SNN, and SkipRNN+SNN on two neuromorphic datasets. 
For N-MNIST dataset, even without a complex architecture, the proposed SkipSNN and their competitors still perform well. 
We find that there is only a slight difference in accuracy between SkipSNN and SNN (\textit{i.e.}, only about $2$\%). 
This is a negligible difference. 
However, as we can observe, there is a significant improvement in the MFLOP count between SkipSNN and SNN. 
Compared to SNN, SkipSNN can provide only half the computational cost of SNN.
Meanwhile, when the computational cost of all models is similar, SkipSNN achieves the higher accuracy than fixed-skip SNN, random-skip SNN and SkipRNN+SNN.

On DVS-Gesture, our proposed model SkipSNN incurs a slight degradation in accuracy (\textit{i.e.}, decrease about $6$\%), but provides a significant speedup -- more than $10x$ times. 
The accuracy of SkipSNN at this level of MFLOP is still higher than other competitors. 
Our experimental results show that SkipSNN using our proposed network structure can greatly speed up inference while only incurring a minimal and negligible loss in classification performance. 
In summary, our proposed model achieves better computational efficiency than previous works when tested on neuromorphic datasets and achieves very negligible degradation in accuracy. 
Moreover, according to Table \ref{tab:nmnist} and \ref{tab:dvs}, our model can also improve the accuracy performance when we consider the trade-off between accuracy and computational cost. 
When SkipSNN considers about $90$\% of updated time-step, it can slightly increase accuracy (\textit{i.e.}, increase about $6$\% on N-MNIST and $0.7$\% on DVS-Gesture). 
This observation is consistent with the conclusion of the first research question. 


\section{Conclusion}
To address spike train classification effectively, a model should focus computational effort only when relevant signals appear. Current SNNs overlook temporal noise, making them computationally intensive and power-hungry. To overcome this, we introduce SkipSNN, a novel model incorporating an event attention mechanism that enables dynamic emphasis on useful signals. SkipSNN reduces computational cost by selectively skipping membrane potential updates, effectively filtering out noise. In tests on neuromorphic datasets N-MNIST and DVS-Gesture, SkipSNN outperforms fixed- and random-skip SNNs, as well as an SNN adaptation of SkipRNN, achieving superior efficiency and accuracy.

\balance
\bibliographystyle{IEEEtran}
\bibliography{reference}

\begin{thebibliography}{10}
\providecommand{\url}[1]{#1}
\csname url@samestyle\endcsname
\providecommand{\newblock}{\relax}
\providecommand{\bibinfo}[2]{#2}
\providecommand{\BIBentrySTDinterwordspacing}{\spaceskip=0pt\relax}
\providecommand{\BIBentryALTinterwordstretchfactor}{4}
\providecommand{\BIBentryALTinterwordspacing}{\spaceskip=\fontdimen2\font plus
\BIBentryALTinterwordstretchfactor\fontdimen3\font minus \fontdimen4\font\relax}
\providecommand{\BIBforeignlanguage}[2]{{%
\expandafter\ifx\csname l@#1\endcsname\relax
\typeout{** WARNING: IEEEtran.bst: No hyphenation pattern has been}%
\typeout{** loaded for the language `#1'. Using the pattern for}%
\typeout{** the default language instead.}%
\else
\language=\csname l@#1\endcsname
\fi
#2}}
\providecommand{\BIBdecl}{\relax}
\BIBdecl

\bibitem{mitra2008real}
S.~Mitra, S.~Fusi, and G.~Indiveri, ``Real-time classification of complex patterns using spike-based learning in neuromorphic vlsi,'' \emph{IEEE TBioCAS}, vol.~3, no.~1, pp. 32--42, 2008.

\bibitem{pfeiffer2018deep}
M.~Pfeiffer and T.~Pfeil, ``Deep learning with spiking neurons: opportunities and challenges,'' \emph{Front. Neurosci.}, vol.~12, p. 774, 2018.

\bibitem{drazen2011toward}
D.~Drazen, P.~Lichtsteiner, P.~H{\"a}fliger, T.~Delbr{\"u}ck, and A.~Jensen, ``Toward real-time particle tracking using an event-based dynamic vision sensor,'' \emph{Experiments in Fluids}, vol.~51, no.~5, pp. 1465--1469, 2011.

\bibitem{mueggler2017event}
E.~Mueggler, H.~Rebecq, G.~Gallego, T.~Delbruck, and D.~Scaramuzza, ``The event-camera dataset and simulator: Event-based data for pose estimation, visual odometry, and slam,'' \emph{IJRR}, vol.~36, no.~2, pp. 142--149, 2017.

\bibitem{rebecq2019high}
H.~Rebecq, R.~Ranftl, V.~Koltun, and D.~Scaramuzza, ``High speed and high dynamic range video with an event camera,'' \emph{IEEE PAMI}, vol.~43, no.~6, pp. 1964--1980, 2019.

\bibitem{wu2019direct}
Y.~Wu, L.~Deng, G.~Li, J.~Zhu, Y.~Xie, and L.~Shi, ``Direct training for spiking neural networks: Faster, larger, better,'' in \emph{AAAI}, vol.~33, 2019, pp. 1311--1318.

\bibitem{yin2021energy}
H.~Yin, J.~B. Lee, X.~Kong, T.~Hartvigsen, and S.~Xie, ``Energy-efficient models for high-dimensional spike train classification using sparse spiking neural networks,'' in \emph{SIGKDD}, 2021, pp. 2017--2025.

\bibitem{wu2018spatio}
Y.~Wu, L.~Deng, G.~Li, J.~Zhu, and L.~Shi, ``Spatio-temporal backpropagation for training high-performance spiking neural networks,'' \emph{Front. Neurosci.}, vol.~12, p. 331, 2018.

\bibitem{shrestha2018slayer}
S.~B. Shrestha and G.~Orchard, ``Slayer: Spike layer error reassignment in time,'' \emph{arXiv preprint arXiv:1810.08646}, 2018.

\bibitem{xin2001supervised}
J.~Xin and M.~J. Embrechts, ``Supervised learning with spiking neural networks,'' in \emph{IJCNN}, vol.~3, 2001, pp. 1772--1777.

\bibitem{schemmel2006implementing}
J.~Schemmel, A.~Grubl, K.~Meier, and E.~Mueller, ``Implementing synaptic plasticity in a vlsi spiking neural network model,'' in \emph{IJCNN}, 2006, pp. 1--6.

\bibitem{wade2010swat}
J.~J. Wade, L.~J. McDaid, J.~A. Santos, and H.~M. Sayers, ``Swat: a spiking neural network training algorithm for classification problems,'' \emph{IEEE Trans. Neural Netw.}, vol.~21, no.~11, pp. 1817--1830, 2010.

\bibitem{gallego2019event}
G.~Gallego, T.~Delbruck, G.~Orchard, C.~Bartolozzi, B.~Taba, A.~Censi, S.~Leutenegger, A.~Davison, J.~Conradt, K.~Daniilidis \emph{et~al.}, ``Event-based vision: A survey,'' \emph{arXiv preprint arXiv:1904.08405}, 2019.

\bibitem{moeys2017sensitive}
D.~P. Moeys, F.~Corradi, C.~Li, S.~A. Bamford, L.~Longinotti, F.~F. Voigt, S.~Berry, G.~Taverni, F.~Helmchen, and T.~Delbruck, ``A sensitive dynamic and active pixel vision sensor for color or neural imaging applications,'' \emph{IEEE TBioCAS}, vol.~12, no.~1, pp. 123--136, 2017.

\bibitem{nozaki2017temperature}
Y.~Nozaki and T.~Delbruck, ``Temperature and parasitic photocurrent effects in dynamic vision sensors,'' \emph{IEEE T-ED}, vol.~64, no.~8, pp. 3239--3245, 2017.

\bibitem{brown2004multiple}
E.~N. Brown, R.~E. Kass, and P.~P. Mitra, ``Multiple neural spike train data analysis: state-of-the-art and future challenges,'' \emph{Nature neuroscience}, vol.~7, no.~5, pp. 456--461, 2004.

\bibitem{glorot2010understanding}
X.~Glorot and Y.~Bengio, ``Understanding the difficulty of training deep feedforward neural networks,'' in \emph{AISTATS}.\hskip 1em plus 0.5em minus 0.4em\relax JMLR Workshop and Conference Proceedings, 2010, pp. 249--256.

\bibitem{campos2017skip}
V.~Campos, B.~Jou, X.~Gir{\'o}-i Nieto, J.~Torres, and S.-F. Chang, ``Skip rnn: Learning to skip state updates in recurrent neural networks,'' \emph{arXiv preprint arXiv:1708.06834}, 2017.

\bibitem{diehl2015fast}
P.~U. Diehl, D.~Neil, J.~Binas, M.~Cook, S.-C. Liu, and M.~Pfeiffer, ``Fast-classifying, high-accuracy spiking deep networks through weight and threshold balancing,'' in \emph{IJCNN}, 2015, pp. 1--8.

\bibitem{cao2015spiking}
Y.~Cao, Y.~Chen, and D.~Khosla, ``Spiking deep convolutional neural networks for energy-efficient object recognition,'' \emph{IJCV}, vol. 113, no.~1, pp. 54--66, 2015.

\bibitem{hu2018spiking}
Y.~Hu, H.~Tang, and G.~Pan, ``Spiking deep residual network,'' \emph{arXiv preprint arXiv:1805.01352}, 2018.

\bibitem{midya2019artificial}
R.~Midya, Z.~Wang, S.~Asapu, S.~Joshi, Y.~Li, Y.~Zhuo, W.~Song, H.~Jiang, N.~Upadhay, M.~Rao \emph{et~al.}, ``Artificial neural network (ann) to spiking neural network (snn) converters based on diffusive memristors,'' \emph{Advanced Electronic Materials}, vol.~5, no.~9, p. 1900060, 2019.

\bibitem{lalapura2021recurrent}
V.~S. Lalapura, J.~Amudha, and H.~S. Satheesh, ``Recurrent neural networks for edge intelligence: A survey,'' \emph{ACM CSUR}, vol.~54, no.~4, pp. 1--38, 2021.

\bibitem{lei2021attention}
T.~Lei, ``When attention meets fast recurrence: Training language models with reduced compute,'' in \emph{EMNLP}, 2021, pp. 7633--7648.

\bibitem{morais2021learning}
R.~Morais, V.~Le, S.~Venkatesh, and T.~Tran, ``Learning asynchronous and sparse human-object interaction in videos,'' in \emph{CVPR}, 2021, pp. 16\,041--16\,050.

\bibitem{zheng2021accurate}
W.~Zheng and G.~Chen, ``An accurate gru-based power time-series prediction approach with selective state updating and stochastic optimization,'' \emph{IEEE Trans. Cybern.}, 2021.

\bibitem{hartvigsen2020learning}
T.~Hartvigsen, C.~Sen, X.~Kong, and E.~Rundensteiner, ``Learning to selectively update state neurons in recurrent networks,'' in \emph{CIKM}, 2020, pp. 485--494.

\bibitem{jernite2016variable}
Y.~Jernite, E.~Grave, A.~Joulin, and T.~Mikolov, ``Variable computation in recurrent neural networks,'' \emph{arXiv preprint arXiv:1611.06188}, 2016.

\bibitem{shen2018ordered}
Y.~Shen, S.~Tan, A.~Sordoni, and A.~Courville, ``Ordered neurons: Integrating tree structures into recurrent neural networks,'' \emph{arXiv preprint arXiv:1810.09536}, 2018.

\bibitem{klimesch2012alpha}
W.~Klimesch, ``Alpha-band oscillations, attention, and controlled access to stored information,'' \emph{TiCS}, vol.~16, no.~12, pp. 606--617, 2012.

\bibitem{buzsaki2006rhythms}
G.~Buzsaki, \emph{Rhythms of the Brain}.\hskip 1em plus 0.5em minus 0.4em\relax Oxford university press, 2006.

\bibitem{orchard2015converting}
G.~Orchard, A.~Jayawant, G.~K. Cohen, and N.~Thakor, ``Converting static image datasets to spiking neuromorphic datasets using saccades,'' \emph{Front. Neurosci.}, vol.~9, p. 437, 2015.

\bibitem{amir2017low}
A.~Amir, B.~Taba, D.~Berg, T.~Melano, J.~McKinstry, C.~Di~Nolfo, T.~Nayak, A.~Andreopoulos, G.~Garreau, M.~Mendoza \emph{et~al.}, ``A low power, fully event-based gesture recognition system,'' in \emph{CVPR}, 2017, pp. 7243--7252.

\end{thebibliography}
\end{document}